\documentclass[letterpaper,twocolumn,10pt]{article}
\usepackage{usenix-2020-09}
\usepackage[T1]{fontenc}
\usepackage[utf8]{inputenc}
\usepackage{microtype}
\usepackage{graphicx}
\usepackage{booktabs}
\usepackage{enumitem}
\usepackage{xcolor}
\usepackage{amsmath}
\usepackage{amssymb}
\usepackage{tikz}
\usetikzlibrary{arrows.meta,positioning,fit,shapes.geometric,shapes.symbols}
\usepackage[hidelinks]{hyperref}
\usepackage{url}
\usepackage{balance}
\usepackage{xspace}
\usepackage{makecell}
\usepackage{multirow}

\newcommand{\sys}{\textsc{KVMem}\xspace}

\title{\textsc{KVMem}: Virtualizing Million-Token Agent Workspaces on a Consumer GPU}
\author{Di Chai$^1$, Leye Wang$^{2*}$, Zeshen Su$^2$, Zhiguo Xia$^{3\#}$, Zhihang Yu$^{4\#}$\\
$^1$\textit{Shanghai University of Finance and Economics} \quad $^2$\textit{Peking University}\\
$^3$\textit{Northwestern Polytechnical University} \quad $^4$\textit{Jilin University}\\
$^*$Corresponding Author \quad $^\#$Work done during an internship at Peking University
}
\date{}

\begin{document}
\maketitle

\begin{abstract}

Modern LLM agents operate in persistent workspaces whose
accumulated history can exceed both GPU KV capacity and the
model's native context window. Existing systems typically compact
older context into summaries or retrieve it later as text, either
losing fine-grained execution evidence or repeatedly prefilling
content that the model has already processed.
We present \sys\footnote{Source code: \url{https://github.com/kvmem/kvmem-qw3}}, a KV-context virtualization system that preserves
overflowed workspace history as paged KV state across GPU memory,
host memory, and NVMe. \sys uses lightweight, model-native
attention-space indexes to select relevant historical blocks and materializes a query-dependent execution view bounded by the model's native context window.
Extensive evaluations on long-context agent benchmarks spanning histories up to one million tokens, including LongMemEval, MemoryAgentBench, and AgentLongBench, show that \sys generally achieves higher task utility and greater inference efficiency than compaction-based approaches, the de facto standard for handling context overflow. In the DeepSWE long-context test with Qwen3.8-27B, \sys improves task success from 43.8\% with compaction-only context management to 48.4\%.

In our local-deployment evaluation, \sys runs Qwen3.6/3.8-27B NVFP4 with MTP on an off-the-shelf laptop equipped with a 24\,GB RTX 5090 Laptop GPU, virtualizing agent workspaces of up to 1M tokens—four times the model's native 256K-token context window. In a single-session setting, \sys generates $\sim$50 tokens/s,
providing interactive responsiveness for local agent execution. More broadly, by decoupling addressable workspace size from the LLM's native context window, \sys provides a practical path toward long-running agents whose workspaces can grow beyond that window.

\end{abstract}

\section{Introduction}
\label{sec:intro}

Modern LLM agents increasingly work inside persistent task workspaces, such as project folders, software worktrees, or notebook directories. A workspace is more than a long conversation transcript. It is the task-local environment in which an agent repeatedly observes state, invokes tools, reads and edits artifacts, incorporates feedback, and accumulates intermediate results. Consequently, a long-running agent builds up a large amount of model-relevant working context. 

This growing workspace context faces two distinct execution limits. First, even when the accumulated history remains within the model’s native context window, retaining all of its KV state on GPU may exceed the memory available to a single agent session. Model weights, runtime buffers, speculative-decoding state, and KV cache compete for the same limited GPU capacity, making the practically usable context substantially smaller than the model’s advertised maximum window on commodity hardware. Second, as execution continues, the accumulated workspace may itself grow beyond the model’s native context window. For example, Qwen3.6-27B supports a native context window of 256K tokens\footnote{\url{https://huggingface.co/Qwen/Qwen3.6-27B}}, whereas long-context agent benchmarks already contain interaction trajectories ranging from hundreds of thousands to several million tokens~\cite{fang2026agentlongbench}. Once either limit is reached, some previously processed context must leave the active model input, even though its exact contents may still be needed later.

Existing agent systems generally handle this overflow by reducing the amount of historical text presented to the model. The most common mechanism is compaction, which replaces older execution history with shorter summaries. Systems may additionally retain the original history in an external text store and retrieve selected snippets when the compact representation is insufficient~\cite{chhikara2025mem0,kang2025memoryos,nan2025nemori,packer2023memgpt}. Although these mechanisms differ in how memory is summarized, indexed, and recalled, they share a text-centric abstraction: once workspace history leaves the active context, it is retained and later recovered primarily as text.

This abstraction creates a coupled fidelity–efficiency problem. Compaction must decide what information to preserve before future agent steps reveal which details will matter. It may therefore omit a sentence from an earlier file, a value from a tool output, or a user constraint that later becomes task-critical. Text retrieval can recover some omitted evidence, but every retrieved token must be processed through prefill again, even though the same content was already processed earlier in the workspace. Text-centric overflow handling not only approximates the historical workspace state, but also discards the reusable KV computation associated with that state.

The loss of reusable KV state, however, is not inherent to context overflow. By the time historical context leaves the active execution window, the model has already encoded it into KV state; the only uncertainty is which subset will be needed again. Drawing inspiration from virtual memory, our key insight is to decouple the size of an agent’s addressable KV workspace from the amount of KV state physically resident on the GPU. Overflowed workspace history can then be preserved as paged KV state in a larger backing store and selectively materialized when relevant. Under this abstraction, the complete KV workspace need not fit in GPU memory at once. Instead, each agent step operates on a bounded, query-dependent execution view drawn from a much larger virtual workspace.

Based on this insight, we present \sys, a KV-context virtualization system for long-running LLM agents. \sys preserves previously processed workspace history as an addressable repository of KV blocks spanning GPU memory, host memory, and NVMe storage. At each agent step, it uses compact attention-space indexes derived from the serving model itself to select relevant historical KV blocks. It then materializes the selected blocks as a chronologically ordered and position-consistent execution view within the model’s native context window and the available GPU KV budget. \sys therefore enables an agent to access a workspace much larger than its active context without reconstructing every recalled block as text and prefilling it again.

Making workspace-context virtualization practical requires addressing three challenges. First, the system must determine when historical context should be reconsidered. Retrieval that occurs too frequently adds ranking and working-set reconstruction overhead, whereas retrieval that occurs too infrequently may leave the model with stale context. Second, the system must determine what to recall from a workspace containing millions of historical tokens in an efficient way without scanning every stored KV tensor. Third, it must determine how to restore a sparse set of non-contiguous KV blocks efficiently and safely. 

\sys addresses these challenges with three corresponding mechanisms. First, \textit{step-level memory scheduling} updates the historical working set once per agent step, around agent-step boundaries where the model's historical attention is most likely to shift. Concretely, the update occurs after prefill and before decoding, and the selected working set remains fixed throughout decoding. Second, \textit{query-conditioned KV retrieval} summarizes each historical block with a compact position-independent representation and ranks blocks using the serving model's own attention-space signals. Third, \textit{tiered KV management} separates the persistent repository view from the GPU execution view. It proactively stages out completed blocks, reuses overlapping GPU pages across consecutive steps, keeps frequently recalled blocks in host memory, and rematerializes the remaining blocks through packed transfers and position-consistent re-RoPE. These mechanisms allow \sys to maintain a large virtual workspace while keeping each model invocation compact and executable on limited GPU memory.

\sys also makes million-token agent workspaces practical on commodity local hardware. On an off-the-shelf laptop equipped with a 24 GB RTX 5090 Laptop GPU, \sys runs Unsloth’s NVFP4-MTP variant of Qwen3.6/3.8-27B and virtualizes a workspace containing up to 1M tokens—four times the model’s native 256K-token context window. In a single-session setting, it sustains $\sim$50 generated tokens per second, providing interactive responsiveness for local agent execution. These results demonstrate that \sys not only reduces the cost of recalling historical context, but also enables substantially larger agent workspaces under limited GPU memory.

In summary, this paper makes three contributions:

1. We identify accumulated agent context as a virtualizable systems resource and distinguish the addressable workspace from both the model-visible execution window and the GPU-resident KV set. This abstraction allows a long-running workspace to exceed the model’s native context window without requiring the model to attend to the entire history in a single invocation.

2. Based on the above abstraction, we present \sys, which preserves overflowed agent context as recoverable KV blocks across GPU memory, host memory, and NVMe. \sys combines step-level memory scheduling, model-native attention-space retrieval, hierarchical KV placement and reuse, and position-consistent rematerialization to construct a bounded execution view at each agent step.

3. We evaluate \sys on three controlled long-context agent benchmarks and show that it generally delivers higher utility and substantially lower recovery overhead than compaction, the de facto standard for handling context overflow in current LLM agent systems. We further evaluate \sys during complete long-horizon software-engineering trajectories, i.e., DeepSWE, with Qwen3.8-27B, and demonstrate a 1M-token virtual workspace on a laptop with only 24 GB of GPU memory while sustaining approximately 50 tokens/s in single-session generation.

\section{Background and Motivation}
\label{sec:motivation}

\subsection{Agent Workspaces and Context Overflow}

An agent workspace is the task-local environment in which an agent reads artifacts, invokes tools, edits files, and accumulates execution state during a run. This pattern is widely used in workspace-centered systems such as SWE-agent~\cite{yang2024sweagent}, Codex~\cite{openai2025codex}, Claude Code~\cite{anthropic2026memory}, OpenHands~\cite{openhands2026workspace}, and OpenClaw~\cite{openclaw2026context}. In these systems, each LLM call constructs a \emph{model-visible context} from workspace instructions, recent dialogue turns, file contents, tool outputs, retrieved notes, and summaries of older history. We focus on long-running agent execution, where workspace memory grows over time and can eventually exceed the model's active context window. This pressure remains practical even for large-window coding agents, since a 200K-token context can be exhausted by repeated file reads, tool traces, logs, edits, and discussions accumulated in a realistic workspace.

A common response to context overflow is \emph{compaction}, which replaces earlier workspace history with a shorter textual representation before execution continues. One widely used implementation is prompt-based summarization, where the runtime asks an LLM to read the prior execution trace and produce a concise handoff summary covering progress, decisions, constraints, and next steps. This pattern appears in compact-style coding-agent workflows~\cite{openai2025codex,anthropic2026contextwindow}. Other systems introduce an explicit context-management component that condenses or prunes the event stream, such as the OpenHands condenser and the OpenClaw compaction/context engine~\cite{openhands2026contextcondenser,openclaw2026compaction}. These mechanisms differ in when compaction is triggered and how much recent context is retained verbatim, but they share the same core abstraction, where overflowed workspace state is compressed into text before being shown again to the main agent.

Retrieval is a natural extension of compaction because a compact summary may omit details that later become relevant. Instead of relying only on the summary, the system can store older memory records or historical snippets outside the active prompt, rank them against the current step, and append selected text back into the model-visible context. This text-centric memory pattern appears in systems such as MemGPT~\cite{packer2023memgpt}, Mem0~\cite{chhikara2025mem0}, MemoryOS~\cite{kang2025memoryos}, and Nemori~\cite{nan2025nemori}, which maintain external memories that can be recalled when a task appears to require them. Applied to agent workspaces, this design yields a simple recovery path, where the system compacts older execution history to stay within the context budget, retrieves raw historical text when the summary is insufficient, and continues with that evidence reinserted into the prompt.

\subsection{Compaction Can Lose Task-Critical Details}

Although compaction keeps a growing workspace within a finite context
window, it can remove information that later becomes necessary for the
task. The core problem is that compaction makes an early decision about
future relevance: it selects what to preserve before later questions,
tool results, or reasoning steps are known. As a result, a summary may
omit low-salience evidence that later becomes essential. This
information-preservation challenge is also observed in prior prompt
compression studies, which show that compressed representations can
fail to retain key details from the original context and consequently
degrade downstream performance~\cite{pan2024llmlingua}.

This limitation can also be understood through a simple
indistinguishability argument. Let $H$ denote the original workspace
history and $C(H)$ its compact representation. Because compaction is
lossy, there can exist two different histories $H_1$ and $H_2$ such
that
\begin{equation}
    C(H_1)=C(H_2), \qquad H_1\neq H_2.
\end{equation}
Consider a future query $q$ whose answer depends on information that
differs between $H_1$ and $H_2$. Once only $C(H)$ remains in the
active context, the model receives the same compact representation for
both histories and therefore cannot recover the omitted distinction
from the summary alone. Thus, unless future relevance is known at
compaction time, a lossy compact representation cannot guarantee that
all information needed by future agent steps is preserved.

For long-running agents, this uncertainty is particularly important:
a sentence in an earlier file, a value in a tool output, or a user
constraint may appear unimportant when compaction occurs but become
critical later. Increasing the summary budget can reduce
the risk of information loss, but also consumes more active-context
capacity and ultimately faces the same bounded-context constraint.
Compaction therefore introduces a fundamental tension between reducing
context size and preserving fine-grained historical evidence.

\subsection{Text Retrieval Recovers Details but Repeats Prefill}

Text retrieval can recover evidence omitted by compaction, but the
retrieved content has already been processed by the model earlier in
the same workspace. Before compaction, a historical token sequence
$x$ has already undergone prefill to produce its KV state,
\begin{equation}
    x \xrightarrow{\mathrm{prefill}} (K_x,V_x).
\end{equation}
If only the text $x$ is retained after compaction, retrieving it at a
later step requires the model to perform this computation again to
reconstruct usable KV state.

More generally, let $R$ denote the historical text retrieved for a
current agent step and let $N_R$ be its number of tokens. A
text-centric recovery path incurs
\begin{equation}
T_{\mathrm{text}}
=
T_{\mathrm{retrieve}}
+
T_{\mathrm{prefill}}(N_R),
\end{equation}
where $T_{\mathrm{prefill}}(N_R)$ is the model computation required to
process the recalled tokens again. Increasing the retrieval budget may
improve the chance of recovering task-relevant evidence, but also
increases this fresh-prefill cost. Prior KV-reuse systems similarly
identify repeated prefill as a major source of inference overhead and
preserve previously computed KV state to avoid recomputing the full
input~\cite{cheng2025lmcache,yao2025cacheblend}.


This observation suggests an alternative recovery path. If the KV
state produced when the workspace history was first processed is
preserved, recall can instead be expressed as
\begin{equation}
T_{\mathrm{KV}}
=
T_{\mathrm{retrieve}}^{\mathrm{KV}}
+
T_{\mathrm{load}}
+
T_{\mathrm{restore}},
\end{equation}
replacing repeated historical prefill with KV loading and restoration.
Whether this path is faster depends on storage, transfer, and
restoration costs.
The key opportunity is that previously computed model state need not be
discarded simply because it leaves the active execution view.

\paragraph{Summary.}
Text-centric overflow handling therefore creates a coupled
fidelity--efficiency problem. Compaction must decide what to preserve
before future relevance is known, while retrieving omitted history as
text requires the model to recompute previously processed content.
\sys targets this gap by preserving overflowed workspace state as
reusable KV state, enabling later recall without turning each memory
access into a new historical-text prefill workload.

\section{Problem Formulation and Challenges}
\label{sec:problem}

\subsection{Problem Formulation}

We study workspace memory for a single long-running agent task.
During execution, the agent accumulates model-visible workspace
history consisting of user instructions, file excerpts, tool outputs,
execution logs, edits, intermediate artifacts, and task constraints.
As this history grows, only a bounded subset can remain in the active
execution view because the view is constrained by both the model's
native context window and the amount of KV state that can physically
reside in GPU memory. Context overflow occurs once useful previously
processed workspace state exceeds either of these limits.

Current agent systems usually handle such overflowed history by
reducing it to compact text, such as summaries or compact memory
records~\cite{anthropic2026contextwindow,openclaw2026compaction,openhands2026contextcondenser}.
We extend this design space by preserving previously processed
workspace history as KV blocks in a larger addressable repository.
Each block preserves the KV state computed when the model processed
the corresponding workspace segment. The system can later materialize
a subset of these historical blocks into a bounded execution view when
they become relevant again.

The goal is to keep a workspace much larger than the active execution
view addressable at high fidelity. Rather than reducing overflowed
context exclusively to compact text, the system uses host memory and
NVMe as backing storage for previously computed KV state and
materializes only a bounded historical working set when needed. A
practical workspace-memory system should recover task-relevant history,
avoid repeatedly prefilling previously processed text, and keep
retrieval and restoration costs within the configured resource
budgets.

More formally, let $H_t$ denote the workspace history that has been
processed by execution point $t$. The KV state of this history is
partitioned into logical blocks of $B_{\mathrm{blk}}$ tokens,
\[
\mathcal{B}_t = \{b_1,b_2,\ldots,b_{M_t}\},
\]
where each logical block is the unit of retrieval scoring, selection,
and KV movement.

We distinguish three capacity limits. Let $B_{\mathrm{model}}$ denote
the model's native context-window size, $B_{\mathrm{gpu}}$ denote the
maximum token-equivalent KV state that can reside under the GPU-memory
budget, and $B_m$ denote the token-equivalent capacity of the backing KV workspace in host memory and NVMe. The effective active-context budget
is therefore
\begin{equation}
B_a = \min(B_{\mathrm{model}}, B_{\mathrm{gpu}}).
\label{eq:active_budget}
\end{equation}
This work focuses on the regime in which the accumulated historical KV
state fits within the configured backing workspace,\footnote{When the backing workspace reaches $B_m$, our current
prototype falls back to compacting older history into text to reclaim
capacity. Since $B_m$ can be much larger than the active KV budget $B_a$ with host memory and NVMe backing, this fallback occurs at a much coarser timescale than conventional compaction triggered by active-context overflow. KV-native reclamation
through block selection, merging, or compaction is an interesting
direction for future work.}
\begin{equation}
B_{\mathrm{blk}} |\mathcal{B}_t| \le B_m.
\label{eq:workspace_capacity}
\end{equation}

A workspace-memory policy $\pi$ determines when the historical working
set should be reconsidered during agent execution. Let
$\mathcal{T}^{\pi}$ denote the resulting set of recall points. At a
recall point $\tau \in \mathcal{T}^{\pi}$, the policy selects a
historical working set
$R_{\tau}^{\pi} \subseteq \mathcal{B}_{\tau}$ to materialize into the
active execution view. Between recall points, the previously
materialized historical working set may be reused without performing
a new workspace-wide selection.

Let $Q_{\tau}$ denote the mandatory context at recall point $\tau$
that is not supplied by the selected historical working set, and let
$N(Q_{\tau})$ denote its token length. The resulting execution view
must satisfy
\begin{equation}
N(Q_{\tau})
+
B_{\mathrm{blk}} |R_{\tau}^{\pi}|
\le B_a.
\label{eq:active_capacity}
\end{equation}
Under these constraints, the workspace-memory policy has two
objectives:
\begin{equation}
\max \mathrm{Fidelity}(\pi),
\qquad
\min \mathrm{RecoveryCost}(\pi).
\label{eq:workspace_objectives}
\end{equation}
$\mathrm{Fidelity}(\pi)$ measures task quality and the correctness of
recovered workspace state, while $\mathrm{RecoveryCost}(\pi)$ measures
the online cost of updating and restoring historical working sets.
A practical policy must therefore determine when the historical
working set should be updated, what historical blocks should be
materialized when an update occurs, and how the selected KV state
should be restored efficiently and faithfully. 

\subsection{Design Challenges}

\textbf{Challenge 1 - When to Recall: Timely Recall Scheduling.} 
An agent repeatedly changes its focus as execution progresses, so the historical context needed by one step may differ substantially from that needed by the next. Reconsidering historical context too frequently adds retrieval and working-set reconstruction overhead, whereas doing so too infrequently can leave the model with a stale execution view after its focus has changed. A practical system must therefore schedule recall at appropriate points in agent execution, balancing timely access to relevant history against the overhead of repeated retrieval.

\smallskip
\noindent\textbf{Challenge 2 - What to Recall: Scalable KV-Block Selection.}
Once recall is triggered, the system must identify which historical blocks are relevant to the current model call. A million-token workspace may contain numerous KV blocks, making it impractical to scan or score their full KV tensors at every recall point. Moreover, relevance should reflect the serving model's own use of historical context rather than relying solely on signals from a separate text retriever. A practical system therefore needs a compact, model-aligned representation of historical KV blocks that supports low-overhead selection over large workspaces.

\smallskip
\noindent\textbf{Challenge 3 - How to Recall: Quality-Preserving KV Restoration.}
After relevant blocks have been selected, their KV state must be restored from host memory or NVMe into the current GPU execution view. This is not a simple load. KV states are substantially larger than their source text, so excessive data movement can offset the prefill savings of KV reuse. More importantly, position-encoded K cannot be directly reused after historical blocks are compacted into new logical positions in the current execution view. A practical system must therefore restore selected blocks efficiently while ensuring that their KV state remains position-consistent and safe for subsequent attention computation.

\begin{figure*}[t]
\centering
\includegraphics[width=0.8\textwidth]{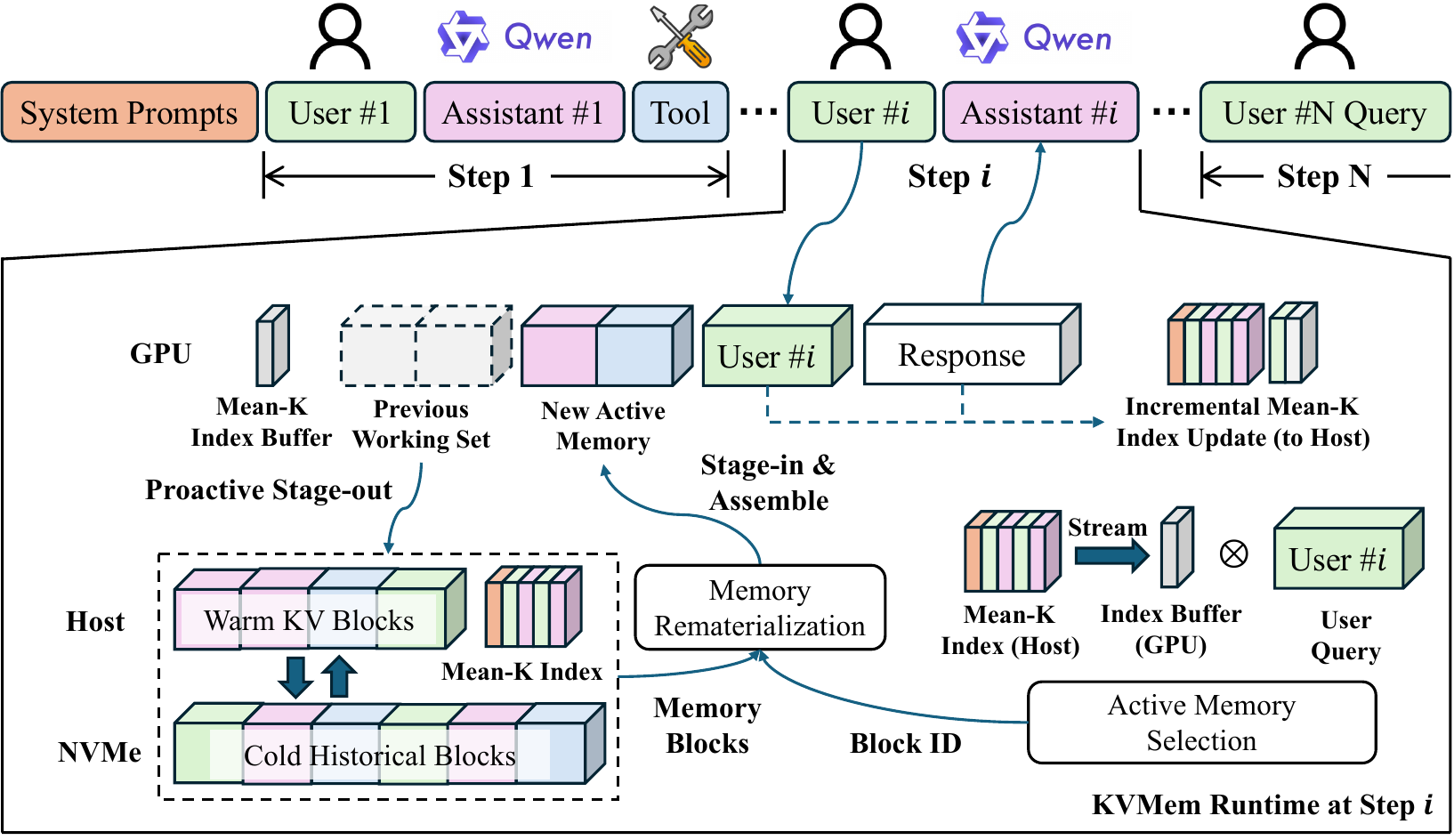}
\caption{Overview of \sys's three core designs. First, step-level memory scheduling reselects a bounded working set once per agent step, after prefill and before decoding (Sec.~\ref{sec-step-level-retrieval}). Second, query-conditioned retrieval represents the full history with compact block-level indexes, enabling efficient block selection with little GPU memory overhead (Sec.~\ref{sec-query-conditioned-retrieval}). Third, tiered KV management proactively stages out completed blocks, reuses overlapping GPU pages, retains frequently retrieved blocks in host memory, and packs the remaining GPU misses into a pipelined rematerialization path (Sec.~\ref{sec-efficient-kv-working-set-management}).}
\label{fig:kvmem-overview}
\end{figure*}

\section{\sys: KV Workspace Virtualization}
\label{sec-design}

\subsection{Overview}
\sys virtualizes an agent's growing workspace by decoupling its addressable KV state from the subset physically resident on the GPU.
Instead of immediately replacing overflowed history with a lossy summary, \sys preserves previously processed workspace state as paged KV blocks across GPU memory, host memory, and NVMe storage. At each agent step, it recalls a bounded, query-dependent subset of historical blocks and materializes them as the active execution view. This allows the addressable workspace to grow far beyond GPU KV capacity while keeping each model invocation within the available GPU-memory and context-window budgets.

Figure~\ref{fig:kvmem-overview} illustrates the architecture and step-level execution flow of \sys. \sys addresses the three recall challenges identified previously with three corresponding mechanisms. 
First, \textbf{step-level memory scheduling} determines \emph{when to recall} historical workspace state, avoiding unnecessary working-set updates when the model's historical focus remains stable (Sec.~\ref{sec-step-level-retrieval}).
Second, \textbf{query-conditioned KV retrieval} determines \emph{what to recall}. It represents historical
blocks using compact, model-native attention-space indexes and ranks them against the current query without scanning their full KV state (Sec.~\ref{sec-query-conditioned-retrieval}). 
Third, \textbf{tiered KV management} determines \emph{how to recall} the selected blocks. It coordinates
page reuse and movement across GPU memory, host memory, and NVMe, and restores selected KV state into a position-consistent GPU execution view (Sec.~\ref{sec-efficient-kv-working-set-management}). 
These mechanisms enable \textit{timely}, \textit{scalable}, and \textit{quality-preserving} recall from a workspace much larger than the GPU-resident working set.

The three mechanisms form a single step-level lifecycle. During the prefill of a new agent step, \sys collects the model-native signals needed for historical-block selection. It then selects and restores a bounded historical working set, combines it with the current context, and replays the current query over the assembled execution view before decoding. The selected historical working set remains fixed during decoding while newly generated KV state grows at its tail. The process repeats in the next agent step, with a new working-set update performed after prefill and before decoding.


\subsection{Step-Level Memory Scheduling}
\label{sec-step-level-retrieval}

Determining when to recall historical workspace state is difficult. Recalling too frequently can track changes in the model's focus closely, but repeatedly scoring the workspace and reconstructing the execution view introduces substantial runtime overhead. Recalling too infrequently, in contrast, can leave the model with a stale historical working set after its focus has shifted. The key scheduling question is therefore when the historical context relevant to the model is likely to change.

\begin{figure}[h]
\centering
\includegraphics[width=\linewidth]{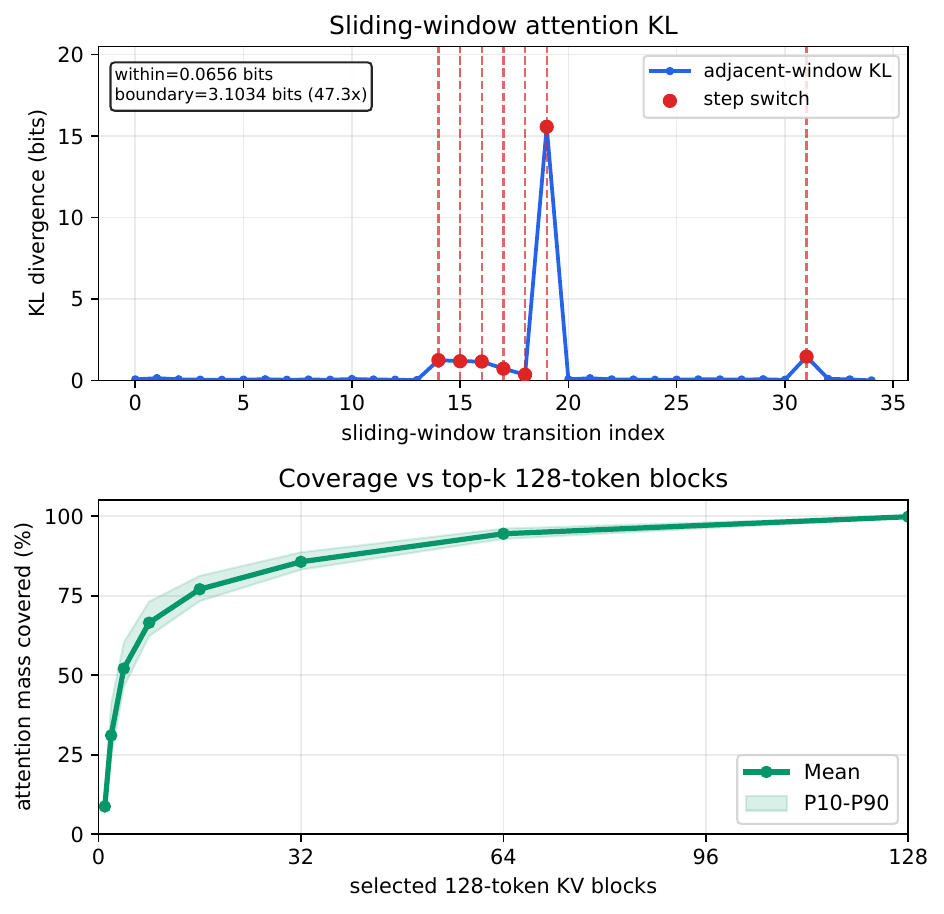}
\caption{Attention behavior across eight OpenHands SWE-bench Lite rollouts. The top panel shows adjacent-window KL for one example rollout, where KL remains low within an agent step and spikes across step boundaries. Red markers and dashed lines denote step switches. The bottom panel reports mean top-$k$ coverage for 128-token KV blocks across all eight rollouts, with the shaded band showing the P10--P90 range.}
\label{fig-attention-step-scheduling}
\end{figure}

To characterize this behavior, we analyze historical attention from eight OpenHands SWE-bench Lite rollouts using adjacent 128-token sliding windows with stride 32. We compare the attention distributions of consecutive windows and measure their KL divergence. Across all eight rollouts, adjacent windows within the same agent step have an average KL divergence of only $0.070$ bits, whereas windows crossing an agent-step boundary average $2.59$ bits, or $37.3\times$ higher. Figure~\ref{fig-attention-step-scheduling} shows a representative rollout, where attention remains stable within each step and changes sharply at step boundaries. These results suggest that historical relevance typically evolves much more slowly within an agent step than across successive steps.


Based on this observation, \sys adopts \emph{step-level memory scheduling}. It updates the historical working set once per agent step, after the current-step prefill and before decoding, and reuses the selected working set throughout the subsequent decode. 
When decoding completes, the next agent invocation begins a new recall epoch in which historical relevance is reconsidered.


Specifically, during the prefill of a new agent step, \sys collects the model-native signals needed for historical selection. Once prefill completes, it performs one working-set update before generating the first decode token. The selected historical set then remains fixed throughout decoding, while newly generated KV state grows at the tail of the execution view. This schedule places recall near observed changes in historical attention while avoiding repeated working-set updates during periods in which the model's historical focus is typically stable. 


\subsection{Query-Conditioned KV Retrieval}
\label{sec-query-conditioned-retrieval}
Once a recall point is reached, \sys must determine which historical KV blocks to retrieve from a workspace that may contain millions of tokens. Scoring the full KV state of every historical token would make both the retrieval index and per-step retrieval cost grow prohibitively with workspace size. At the same time, conventional text retrieval based on lexical features or embeddings from a separate model may not reflect which historical states the serving model itself would attend to. \sys therefore performs scalable, model-native KV retrieval using compact representations derived directly from the serving model's attention space.

Historical attention is also highly sparse. As shown in the bottom panel of Figure~\ref{fig-attention-step-scheduling}, each analyzed window contains 103.1 historical blocks on average, yet the top-8 blocks capture 66.5\% of historical attention mass, while the top-16 capture 77.0\%. This suggests that effective recall does not require materializing most of the workspace; instead, the retrieval mechanism should efficiently identify a small set of strongly relevant historical blocks.


Existing KV-cache retention methods commonly operate at token granularity. H$_2$O ranks individual tokens using their accumulated attention scores~\cite{zhang2023h2o}, while StreamingLLM retains token-level attention sinks and recent tokens~\cite{xiao2024streamingllm}. Although such mechanisms are effective for pruning an active context, directly extending token-level scoring to a persistent million-token workspace would require maintaining and comparing retrieval state for every historical token. This is impractical in both index capacity and query-time latency.

To make model-native KV retrieval scalable, \sys turns token-level KV retrieval into compact block-level operations. During prefill, it summarizes each logical KV block with a position-independent Mean-K representation for every layer and KV head. At each retrieval point, it compares the current query with these Mean-K vectors and scores historical blocks directly in the serving model's attention space. The key insight is that, with sufficiently fine-grained blocks, a simple Mean-K representation can preserve useful block-level relevance while substantially reducing the index size and retrieval cost relative to token-level scoring.

\smallskip

\smallskip
\noindent\textbf{Mean-K Retrieval-Index Construction.}
For each historical KV block, \sys constructs a compact retrieval
representation directly from the serving model's K vectors.
Historical K vectors contain positional information introduced by RoPE
when they are originally computed. Before constructing the retrieval
index, \sys removes this positional encoding so that the resulting K
vectors represent content independently of their original
execution-window positions. The same transformation is applied to the
current query vectors during retrieval, allowing historical K and
current queries to be compared consistently even after blocks are
remapped to new logical positions.

For each layer and KV head, \sys represents a historical block $b$ by
the mean of its position-independent K vectors. Let
$\tilde{k}_{l,i,g}$ denote the position-independent K vector of token
$i$ at layer $l$ and KV head $g$. The Mean-K representation of block
$b$ is
\begin{equation}
\bar{k}_{l,b,g}
=
\frac{1}{|b|}
\sum_{i \in b}
\tilde{k}_{l,i,g}.
\label{eq:mean_k_index}
\end{equation}

This representation stores only one K vector per block, layer, and KV
head, making the retrieval index substantially smaller than the full
historical KV state. Mean-K inevitably loses within-block variation,
but this approximation becomes less severe as the block size decreases,
because each mean summarizes a shorter span with less heterogeneous
content. Smaller blocks also provide finer-grained retrieval and
working-set construction, allowing \sys to select relevant historical
state at higher resolution.
Although smaller blocks increase index and block-management overhead,
\sys remains efficient at fine granularity. Our implementation defaults to 32-token blocks with Mean-K, which provides a practical balance between retrieval fidelity, granularity, and system efficiency without requiring a more complex multi-vector index.

\smallskip
\textbf{Query-Conditioned Block Scoring.}
At each recall point, \sys scores every retrievable historical block
against the current query using its Mean-K representation. For each
layer and query head, the position-independent query vectors are
compared with the Mean-K vectors of all candidate blocks through
scaled dot products. \sys then applies a global softmax across
candidate blocks so that their relevance is normalized under a common
attention-space scale. The configured sink and recent blocks are
always retained.

We use $l$, $m$, and $h$ to index the scored attention layers,
current query tokens, and query heads, respectively. Let
$q_{l,m,h}$ denote the position-independent query vector, let $g(h)$
denote the KV head associated with query head $h$ under grouped-query
attention, and let $\bar{k}_{l,b,g(h)}$ denote the Mean-K vector of
historical block $b$. The head dimension is $d$. Given the candidate
block set $\mathcal{C}$, \sys computes the block relevance as
\begin{equation}
R_b =
\sum_m
\operatorname{mean}_{l,h}
\operatorname{softmax}_{b' \in \mathcal{C}}
\left(
\frac{
q_{l,m,h}^{\top}
\bar{k}_{l,b',g(h)}
}{
\sqrt{d}
}
\right)_b .
\label{eq-kvmem-block-score}
\end{equation}
For every fixed $(l,m,h)$, the softmax normalizes the scaled
dot-product scores across all candidate blocks. The outer operations
sum these relevance scores across query tokens and average them across
the scored layer--head pairs. The default selector first reserves the
configured sink and recent regions and fills the remaining budget with
the blocks having the largest $R_b$ scores. Finally, \sys restores
chronological order before materializing the selected blocks.

Because each block is represented by only one Mean-K vector per layer
and KV head, \sys scores compact block-level representations rather
than processing every historical token individually. For very large
workspaces, the complete Mean-K index is kept in host memory and scored
through bounded tiles staged on the GPU.

\subsection{Tiered KV Management}
\label{sec-efficient-kv-working-set-management}

Once relevant historical blocks have been retrieved, \sys must restore their KV state into the bounded GPU execution view. This restoration is not a simple page load. Historical blocks may reside across GPU memory, host memory, and NVMe storage, and selected blocks must be assembled under a fixed GPU KV budget. Moreover, their position-encoded K may no longer be valid after the blocks are assigned new logical positions in the execution view. \sys therefore needs to coordinate KV storage, reuse, data movement, and positional restoration while keeping the recall overhead low.


Existing inference systems provide several useful but distinct building blocks. PagedAttention-based engines such as vLLM provide efficient paged KV allocation and sharing for active requests, but their paging abstraction does not maintain a persistent, query-dependent workspace across GPU memory, host memory, and NVMe~\cite{kwon2023vllm}. LMCache extends KV reuse beyond GPU memory with persistent tiered storage and optimized data movement, but it does not construct a changing sparse execution view by repeatedly selecting historical workspace blocks and remapping them into new compact positions~\cite{cheng2025lmcache}. CacheBlend further enables non-prefix KV reuse by selectively recomputing cached context to repair attention mismatches, but focuses on fusing already selected cached content rather than managing the lifecycle and retrieval of a continuously growing agent workspace~\cite{yao2025cacheblend}.

\sys builds on these complementary capabilities but targets a different execution model: at every agent step, it must recall a changing, query-dependent subset from a persistent workspace, preserve physical reuse across consecutive working sets, coordinate placement across GPU memory, host memory, and NVMe, and restore the selected blocks into new compact logical positions before decoding.

To support this execution model, \sys separates workspace KV into two coordinated views. The \emph{repository view} tracks each historical block and the locations of its valid copies across a bounded GPU page pool, host memory, and NVMe. The \emph{execution view} contains only the blocks selected for the current step, ordered chronologically and assigned contiguous positions in a compact GPU attention window. The GPU pool reserves capacity for the execution KV set and newly generated tokens. Once the selected blocks are resident, \sys constructs the execution view by mapping their non-contiguous physical pages into the logical order of the compact attention window through a page table rather than copying them into a dense cache. Page-table aliasing alone, however, does not make K valid at its new compact positions. Because RoPE-encoded K is valid only at the positions for which it was encoded, \sys retains an immutable, position-independent raw K outside the active GPU cache as the source of truth. For an incoming block, \sys applies RoPE to raw K at the block's assigned compact position and materializes the result in its destination GPU page, while V is copied unchanged. A block that is already GPU-resident can instead reuse its current K through a bounded delta re-RoPE when its compact position changes. Outside the active GPU cache, raw K remains the reconstruction authority, while lower-tier block records preserve V. Each active GPU page contains K encoded for its current compact position and an unchanged copy of V. Periodically reconstructing K from raw K bounds the numerical error accumulated through repeated in-place position adjustments on a low-precision GPU copy.

The view separation provides the basic virtualization mechanism, but a naive implementation still incurs substantial restoration overhead. It may synchronously stage out outgoing blocks before their GPU pages can be reused, redundantly reload blocks shared by consecutive working sets, and issue many small transfers and reconstruction operations for physically scattered KV pages. \sys addresses these costs with three complementary optimizations: \emph{proactive stage-out}, \emph{retrieval-aware hierarchical reuse}, and \emph{packed and pipelined KV rematerialization}.

\smallskip
\noindent \textbf{Proactive Stage-Out.}
Under a fixed GPU KV budget, outgoing and incoming working-set blocks may not fit on the GPU simultaneously. A naive restoration path must therefore stage out outgoing blocks before reclaiming their pages for newly retrieved blocks, placing stage-out directly on the restoration
critical path.

\sys removes this dependency from the working-set transition by performing stage-out earlier and overlapping it with chunked-prefill computation. After prefill chunk $n$ completes, \sys batches the KV state of completed blocks and asynchronously creates host-memory copies while the GPU processes chunk $n+1$, with background workers continuing persistence to NVMe when necessary.
When the next working-set transition occurs, the outgoing blocks already have valid copies in host memory or NVMe. \sys can then immediately reclaim their GPU pages and stage in the newly selected blocks without waiting for additional stage-out operations. Reclaiming a GPU page removes only its temporary GPU copy, while the corresponding historical block remains recoverable from a lower tier.

\smallskip
\noindent\textbf{Retrieval-Aware Hierarchical Reuse.}
Consecutive agent steps often retrieve overlapping historical blocks. \sys exploits this temporal locality at two levels. First, blocks that remain selected across consecutive working sets can reuse their GPU-resident pages, avoiding redundant H2D transfers. Second, blocks that are repeatedly retrieved but no longer fit on the GPU are preferentially retained in host memory, reducing expensive reloads from NVMe.

To reuse GPU-resident blocks, \sys updates the working set as a delta. Let $W_{t-1}$ and $W_t$ be the working sets selected for two consecutive steps. \sys decomposes the transition into
\begin{align}
  R_t &= W_t \cap W_{t-1}, &
  L_t &= W_t \setminus W_{t-1}, &
  E_t &= W_{t-1} \setminus W_t ,
  \label{eq-kvmem-working-set-delta}
\end{align}
where $R_t$, $L_t$, and $E_t$ are the retained, incoming, and outgoing blocks, respectively. 

More generally, let $G_{t-1}$ denote all historical blocks that remain
resident in the bounded GPU page pool. The blocks that physically
require stage-in are therefore
\begin{equation}
  L_t^{\mathrm{physical}} = W_t \setminus G_{t-1}.
  \label{eq-kvmem-physical-load-set}
\end{equation}
\sys preserves the physical pages of GPU hits and loads only $L_t^{\mathrm{physical}}$. Because proactive stage-out has already
created valid lower-tier copies of outgoing blocks, their GPU pages can
be reclaimed immediately when necessary.

Reusing a GPU-resident page avoids data movement, but its K may no
longer correspond to the block's new logical position. V remains
unchanged, whereas K must be re-RoPEd when the compact position
changes. To avoid numerical error from repeated in-place adjustment on
the low-precision GPU copy of K, \sys bounds the number of delta re-RoPE operations
performed between reconstructions. After the configured limit is
reached, \sys reconstructs K from its position-independent raw
representation, applies RoPE once at the new position, and resets the
counter.

GPU reuse does not help once a block has been evicted from the GPU.
Such blocks remain available in either host memory or NVMe, but
reloading from host memory is substantially faster than reading from
NVMe. Because host-memory capacity is limited, \sys prioritizes blocks
that are likely to be retrieved again. Specifically, host-memory admission and eviction consider both recent
retrieval and cumulative retrieval frequency. Blocks with stronger
recent or long-term reuse are preferentially retained in host memory,
while less frequently retrieved blocks are demoted to NVMe. Compared
with a recency-only policy such as LRU, this policy avoids repeatedly
sending frequently recalled workspace blocks to NVMe after temporary
periods of inactivity.

Together, hierarchical reuse minimizes data movement across both
GPU--host and host--NVMe boundaries without changing the retrieved
working set. Blocks that still miss in the GPU pool are handled by the
packed rematerialization path described next.

\smallskip
\noindent\textbf{Packed and Pipelined KV Rematerialization.}
KV rematerialization converts lower-tier KV state into position-consistent, attention-ready GPU pages. For the remaining GPU misses, however, both the source blocks in host memory or NVMe and their destination pages in the bounded GPU pool are typically non-contiguous. Rematerializing each block independently would therefore require many small data transfers and reconstruction-kernel launches, underutilizing PCIe bandwidth and amplifying launch and synchronization overhead. \sys addresses this problem by decoupling the physical placement of KV pages from the granularity at which they are transferred and reconstructed.

\emph{Packed rematerialization} batches multiple selected blocks into
contiguous transfers. Persistent host workers gather the raw K of
selected blocks into pinned-memory buffers, while adjacent NVMe
extents are coalesced whenever possible. The packed data are then
transferred to bounded GPU staging buffers in bulk. A GPU kernel
scatters the K into their destination KV pages and applies RoPE
according to the newly assigned positions in the compact execution
view. V is position-independent and can be transferred without
positional reconstruction. We store each block's raw K contiguously in host memory to support efficient block-wise gathering, and use precomputed RoPE sine/cosine tables to reduce GPU reconstruction overhead.

\begin{figure}[t]
\centering
\includegraphics[width=\linewidth]{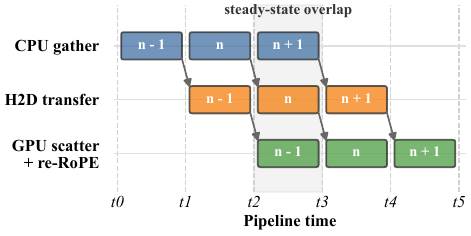}
\caption{Packed and pipelined KV rematerialization. Fragmented historical KV turns stage-in into many small H2D transfers and reconstruction launches, underutilizing PCIe bandwidth and amplifying launch and synchronization overhead. \sys gathers the scattered blocks into contiguous host buffers, transfers them in bulk, and overlaps CPU gather of batch $n+1$, H2D transfer of batch $n$, and GPU scatter/re-RoPE of batch $n-1$.}
\label{fig:kvmem-rematerialization-pipeline}
\end{figure}

\emph{Pipelined rematerialization} overlaps the processing of consecutive packed batches. Each batch passes through three stages: (1) the host gathers selected raw-K blocks into a contiguous pinned-memory buffer, (2) the packed data are transferred to a GPU staging buffer through H2D copy, and (3) the GPU scatters the data into their destination KV pages and applies re-RoPE to K. These stages primarily use different resources---CPU memory bandwidth, the GPU copy engine, and GPU compute---and can therefore execute concurrently for different batches. As illustrated in Figure~\ref{fig:kvmem-rematerialization-pipeline}, while batch $n$ is being transferred to the GPU, the host gathers batch $n+1$, and the GPU simultaneously scatters and re-RoPEs batch $n-1$. Double buffering provides separate host and GPU staging buffers for adjacent batches, allowing these operations to overlap without overwriting data that are still in use. Asynchronous event-based synchronization ensures that a buffer is reused only after its preceding transfer or reconstruction has completed, avoiding device-wide synchronization; our implementation realizes this mechanism using CUDA events. Because V requires no position-dependent transformation, V can be staged in concurrently with raw-K gathering and reconstruction.

Briefly, packing converts fragmented KV restoration into bulk data movement, while pipelining hides much of the remaining transfer and reconstruction latency behind adjacent stages, enabling efficient rematerialization even at fine block granularity.

\section{Implementation}
\label{sec:implementation}

\sys requires end-to-end control over model execution, KV page allocation, positional encoding, and storage transfers. Because these mechanisms span interfaces that existing serving systems do not expose together, we implement \textsc{QW3}, a native inference engine that provides the execution substrate for \sys.

\subsection{QW3 Inference Engine}

Modern inference systems solve several related but narrower problems. vLLM and SGLang provide paged KV allocation and prefix reuse, while LMCache preserves and transfers KV state across storage tiers~\cite{kwon2023vllm,zheng2024sglang,cheng2025lmcache}. Their standard execution abstractions nevertheless present attention with a request prefix whose logical positions grow monotonically, and cached KV pages are restored at those positions. \sys instead constructs a changing, query-dependent execution view at every agent step by recalling non-contiguous historical blocks and remapping them into compact logical positions.
Retrieval further requires access to intermediate Q and K vectors during prefill so that blocks remain searchable after their full KV tensors leave GPU memory. Supporting this execution model in an existing engine would require coordinated modifications to the model executor, attention kernels, page tables, memory allocator, and request scheduler.

We implement \textsc{QW3} as a C++ and CUDA inference engine for Qwen models. Its native executor controls the model forward path, paged KV allocator, attention page tables, CUDA kernels, and device-transfer streams within one runtime. \textsc{QW3} also supports continuous batching, prefix reuse, and multi-token prediction (MTP) for speculative decoding. \sys adds a host-side block manager, retrieval and re-RoPE kernels, and pinned-memory and NVMe storage backends to this executor. It manages the growing KV state of standard RoPE attention layers, while recurrent model state continues through the original execution path.

\subsection{\sys Execution in \textsc{QW3}}

\textsc{QW3} partitions workspace history into configurable logical KV blocks aligned with physical KV-page boundaries. A logical block is \sys's unit of retrieval scoring, selection, and movement, while physical pages are the allocation units managed by \textsc{QW3}.
This alignment allows \sys to move, evict, and remap a logical block without splitting physical pages across different execution regions. Each block records its chronological identifier, original token range, residency tier, lower-tier location, and \textit{baked position}, which denotes the RoPE frame currently encoded in its K. Newly registered blocks initially use their original positional frame.

During prefill, \textsc{QW3} removes RoPE from the K vectors of completed logical KV blocks and computes their Mean-K retrieval representations while the corresponding K vectors are still resident on the GPU. A straightforward implementation would keep the complete attention-space index on the GPU, allowing all historical blocks to be scored directly. However, the index grows linearly with workspace size and can consume substantial GPU memory at multi-million-token scale, undermining the bounded GPU-memory goal of \sys.

\textsc{QW3} adopts a memory-efficient tiled retrieval design. The complete attention-space index is stored in host memory, while only a fixed-size tile is staged on the GPU at a time. At each recall point, \textsc{QW3} captures the current query vectors in the same position-independent space and streams successive index tiles through a bounded GPU staging buffer. Each tile is scored on the GPU, and the results are progressively combined to obtain block-level relevance scores over the complete historical index. \textsc{QW3} maintains the global softmax normalization across tiles, so tiled scoring is equivalent to evaluating Eq.~\eqref{eq-kvmem-block-score} over the complete candidate index. The resulting scores are then used by the host selector, together with the always-retained sink and recent blocks and the active-context budget, to determine the final retrieved block set. This design retains GPU-accelerated retrieval while keeping its GPU-memory footprint bounded as both the workspace and retrieval index grow.

The block manager computes the difference between the previous and next working sets and emits a plan containing stage-out, stage-in, and positional-remapping operations. Selected blocks are ordered chronologically and assigned contiguous positions in the compact window. The data plane then fetches only the selected KV pages, constructs an attention page table over their GPU locations, and transforms their K into the assigned positional frame. 
Because the initial query prefill is used to drive retrieval before the new working set is assembled, \textsc{QW3} re-prefills the current query over the updated execution view before decoding, ensuring that its model state reflects the recalled historical context. Note that \textsc{QW3} avoids replaying earlier context by restoring a checkpoint taken immediately before the current query.
Decode subsequently extends this compact execution view at its tail, while the complete workspace remains available through the repository view.

\subsection{Implementation Optimizations}

\textbf{Incremental Indexing.}
\textsc{QW3} constructs each block's Mean-K retrieval representation during prefill while its K vectors are already resident on the GPU, and subsequently stores the compact index entry in host memory. This avoids an additional scan of historical KV state.

\smallskip

\noindent \textbf{Host-Resident, GPU-Tiled Retrieval Scoring.}
\textsc{QW3} stores the complete index in host memory and streams fixed-size tiles through a bounded GPU staging buffer for scoring. Each tile is scored on the GPU and the partial results are progressively combined across tiles. This preserves GPU-accelerated retrieval while keeping the GPU memory used for retrieval independent of the total index size.

\smallskip

\noindent \textbf{Overlapped tier movement.}
Host-resident KV blocks use pinned memory for asynchronous GPU transfer,
while NVMe I/O is handled by background workers. Data movement is
scheduled on independent transfer streams to overlap with model
execution whenever possible.

\smallskip

\noindent \textbf{Copy-free execution-view assembly.}
\textsc{QW3} assembles the execution view by constructing a compact page table
over the physical GPU pages of the retrieved blocks, avoiding an
additional copy into a dense KV cache. Position changes are handled
through batched re-RoPE, while newly generated KV pages are appended
directly to the same execution view.

\section{Evaluation}
\label{sec:evaluation}

\subsection{Experimental Setup}

We evaluate \sys under two complementary model configurations. Sections \ref{sec_exp:utility}–\ref{sec_exp:scalability} use Qwen3.6-27B for controlled evaluations of task utility, recovery efficiency, consumer-device deployment, and workspace scalability. Section~\ref{sec:deepswe} separately evaluates the same \sys workspace-memory design with the newer Qwen3.8-27B on complete long-horizon software-engineering trajectories, as Qwen3.8 has significantly improved the model's long agent running capability. Note that the Qwen3.8-27B experiment requires no change to \sys's workspace-memory mechanisms, providing an additional test of whether the design transfers to a newer model generation. \sys is implemented in QW3, our native C++ and CUDA inference engine described in Section~\ref{sec:implementation}.

We use two complementary experimental environments.

\textbf{The server platform} is equipped with a 96\,GB NVIDIA RTX PRO 6000 GPU, 128\,GB of host memory, and 4\,TB of NVMe
storage. We run Unsloth's Qwen3.6/3.8-27B with Q8-quantized weights and an FP8 KV cache. The platform can natively accommodate the model's full 256K-token context window, enabling direct comparisons with Full Context whenever the complete history fits within the model window.
By paging overflowed KV state to host memory and NVMe storage, \sys supports an addressable workspace of up to 10M tokens. We use this platform for controlled evaluations of task utility, recovery efficiency, and workspace scalability.

\textbf{The consumer platform} is an off-the-shelf laptop equipped with a 24\,GB NVIDIA RTX 5090 Laptop GPU, 32\,GB of host memory, and 1\,TB of NVMe storage. We run Unsloth's NVFP4-MTP variant of Qwen3.6/3.8-27B with an FP8 KV cache. We use this platform to evaluate the practicality of supporting a 1M-token virtual workspace under consumer-grade resource constraints.

We evaluate on controlled long-history benchmarks where the same interaction history or workspace trace can be replayed across memory policies:

\textbf{- LongMemEval-S}~\cite{wu2025longmemeval}: Given a timestamped multi-session interaction history and a user question, the model must predict the correct answer by recovering evidence from earlier sessions. LongMemEval-S contains roughly 115K tokens per question. 

\textbf{- MemoryAgentBench}~\cite{hu2025memoryagentbench}: Given an incrementally accumulated multi-turn history, the model must answer memory-dependent queries across accurate retrieval, test-time learning, long-range understanding, and conflict resolution. The benchmark covers histories from roughly 103K to 1.44M tokens.

\textbf{- AgentLongBench}~\cite{fang2026agentlongbench}: Given a long simulated agent-environment rollout, the model must solve the final task by synthesizing observations, feedback, and intermediate state accumulated throughout the rollout. AgentLongBench evaluates long-context agent trajectories ranging from 32K to 4M tokens.

We compare \sys with the following baselines:

\textbf{- Full Context}: the complete history is kept within the active context and serves as a reference setting.

\textbf{- Sliding Window}: only the most recent context within the
active-context budget is retained, following recent-window baselines
commonly used in streaming and KV-cache management
~\cite{xiao2024streamingllm,zhang2023h2o}.

\textbf{- Compact-only}: overflowed history is compressed into a textual summary, following compaction-style agent workflows~\cite{openai2025codex,anthropic2026contextwindow,openhands2026contextcondenser,openclaw2026compaction}.

\textbf{- Compact+RAG}: the prompt is augmented with retrieved historical text blocks, following text-memory and retrieval-based agent memory systems~\cite{packer2023memgpt,chhikara2025mem0}.

Within each benchmark setting, Sliding Window, Compact-only, Compact+RAG, and \sys use the same active execution-view budget to ensure a fair comparison. We adjust this budget according to workspace scale: 32K tokens for LongMemEval-S and AgentLongBench ($\leq$256K), 64K tokens for MemoryAgentBench ($>256K$) and AgentLongBench (512K), and 100K tokens for AgentLongBench (1M). Full Context uses the complete history and is reported only when the history fits within the model's native 256K-token context window.  All compared methods are executed with \textsc{QW3}; they differ only in their context-management policies. Unless otherwise specified, \sys uses 32-token logical blocks and Mean-K retrieval throughout the evaluation.

\begin{table*}[t]
\centering
\small
\setlength{\tabcolsep}{6pt}
\renewcommand{\arraystretch}{1.10}

\caption{
Utility and efficiency results across controlled long-history benchmarks.
The active context column reports the context budget used by Sliding Window, Compact-only, Compact+RAG, and \sys; Full Context is not subject to this limit.
Pre-answer latency includes memory construction and retrieval when applicable, together with final-query input processing.
For Compact+RAG, values in parentheses exclude the time
required to generate the compact summary.
All methods are executed with the same \textsc{QW3} backend and model configuration.
}
\label{tab:utility-efficiency}

\begin{tabular}{@{}lclccccc@{}}
\toprule
\makecell[c]{\textbf{Benchmark}}
& \makecell[c]{\textbf{Active}\\\textbf{Context}}
& \makecell[c]{\textbf{Metric}}
& \makecell[c]{\textbf{Full Context}\\\textbf{(ref.)}}
& \makecell[c]{\textbf{Sliding}\\\textbf{Window}}
& \makecell[c]{\textbf{Compact}\\\textbf{Only}}
& \makecell[c]{\textbf{Compact}\\\textbf{+RAG}}
& \makecell[c]{\textbf{\sys}} \\
\midrule

\multirow{2}{*}{LongMemEval-S}
& \multirow{2}{*}{32K}
& Answer Acc. (\%) $\uparrow$
& 86.60
& 26.80
& 45.60
& \textbf{86.20}
& \underline{85.60} \\

&
& Latency (s) $\downarrow$
& 0.30
& \textbf{0.19}
& 18.92
& 26.63 (10.63)
& \underline{0.48} \\
\midrule

\multirow{2}{*}{\shortstack[l]{MemoryAgentBench\\($>256$K)}}
& \multirow{2}{*}{64K}
& Overall Score (\%) $\uparrow$
& --
& 17.95
& 27.54
& \underline{34.80}
& \textbf{40.99} \\

&
& Latency (s) $\downarrow$
& --
& \textbf{0.26}
& 85.87
& 106.02 (20.72)
& \underline{1.81} \\
\midrule

\multirow{2}{*}{\shortstack[l]{AgentLongBench\\($\leq$256K)}}
& \multirow{2}{*}{32K}
& Task Success (\%) $\uparrow$
& 59.54
& 25.36
& 15.84
& \underline{47.49}
& \textbf{60.87} \\

&
& Latency (s) $\downarrow$
& 0.17
& \textbf{0.11}
& 97.04
& 111.39 (14.80)
& \underline{0.38} \\
\midrule

\multirow{2}{*}{\shortstack[l]{AgentLongBench\\(512K)}}
& \multirow{2}{*}{64K}
& Task Success (\%) $\uparrow$
& --
& 25.00
& 22.50
& \textbf{54.00}
& \underline{53.00} \\

&
& Latency (s) $\downarrow$
& --
& \textbf{0.20}
& 246.49
& 263.70 (23.58)
& \underline{0.62} \\
\midrule

\multirow{2}{*}{\shortstack[l]{AgentLongBench\\(1M)}}
& \multirow{2}{*}{100K}
& Task Success (\%) $\uparrow$
& --
& 20.00
& 32.00
& \underline{42.00}
& \textbf{50.00} \\

&
& Latency (s) $\downarrow$
& --
& \textbf{0.26}
& 380.19
& 416.38 (39.24)
& \underline{0.73} \\

\bottomrule
\end{tabular}
\end{table*}

\begin{table*}[t]
\centering
\small
\setlength{\tabcolsep}{7pt}
\renewcommand{\arraystretch}{1.12}
\caption{Utility and efficiency results on LongMemEval-S. Higher answer accuracy is better; lower values are better for all efficiency metrics. Full Context assumes sufficient context and KV-cache capacity to include the complete history; all other methods use a 32K-token active context window. Total input includes the input used to construct the compact summary for Compact-only and Compact+RAG. Fresh prefill denotes the total cache-miss input. Pre-answer latency includes memory construction and retrieval when applicable, together with final-query input processing.}
\label{tab:longmemeval-utility-efficiency}

\begin{tabular}{@{}lcccc@{}}
\toprule
& \multicolumn{1}{c}{\textbf{Utility}}
& \multicolumn{3}{c}{\textbf{Efficiency}} \\
\cmidrule(lr){2-2}
\cmidrule(lr){3-5}

\textbf{Method}
& \makecell{\textbf{Answer Acc.} \textbf{(\%) $\uparrow$}}
& \makecell{\textbf{Total} \textbf{input (K) $\downarrow$}}
& \makecell{\textbf{Fresh} \textbf{prefill (K) $\downarrow$}}
& \makecell{\textbf{Pre-answer} \textbf{latency (s) $\downarrow$}} \\
\midrule

Full Context
& 86.60
& 109.74
& 0.08
& 0.30 \\
\midrule

Sliding Window
& 26.80
& \textbf{27.99}
& \textbf{0.08}
& \textbf{0.19} \\

Compact-only
& 45.60
& 114.00
& 2.83
& 18.92 \\

Compact+RAG
& \textbf{86.20}
& 142.16
& 31.00
& 26.63 \\

\sys
& \underline{85.60}
& \underline{109.74}
& \textbf{0.08}
& \underline{0.48} \\

\bottomrule
\end{tabular}
\end{table*}

\subsection{Utility}
\label{sec_exp:utility}

We first evaluate whether \sys preserves task utility when accumulated
workspace history exceeds the active execution budget. Table~\ref{tab:utility-efficiency}
summarizes the results across the three long-history benchmarks.
Overall, \sys consistently outperforms Sliding Window and Compact-only,
showing that preserving and selectively recalling previously processed
workspace state retains substantially more task-relevant information
than either discarding old context or compressing it into summaries.
Compared with Compact+RAG, \sys generally matches or exceeds its utility
without reconstructing recalled history from text.

On LongMemEval-S, where the complete history remains within the
model's native context window and Full Context is therefore available
as a reference, \sys achieves 85.6\% answer accuracy, within
1.0 percentage point of Full Context (86.6\%) and 0.6 points of
Compact+RAG (86.2\%). In contrast, Sliding Window and Compact-only
achieve only 26.8\% and 45.6\%, respectively. This result shows that
\sys can preserve nearly all of the utility of the complete history
while exposing only a bounded, query-dependent execution view.

AgentLongBench exhibits an even stronger result for trajectories within
the native 256K-token context window. \sys achieves 60.87\% task
success, the highest among all evaluated methods, compared with
59.54\% for Full Context and 47.49\% for Compact+RAG. Sliding Window
and Compact-only achieve only 25.36\% and 15.84\%, respectively.
Thus, selectively recalling historical workspace state does not
necessarily sacrifice utility relative to presenting the entire
trajectory; on this benchmark, \sys slightly surpasses the Full
Context reference. One possible explanation is that nominal context-window capacity does not guarantee equally effective use of the entire context: prior work has shown that LLM performance can degrade as context grows, even within supported context lengths~\cite{hsieh2024ruler,liu2024lost}. By exposing only a query-relevant subset of the history, \sys may partially mitigate such long-context degradation.

The need for workspace virtualization becomes particularly clear once the workspace exceeds the model's native 256K-token context window, where Full Context is no longer feasible and \sys must recover task-relevant history from a much larger addressable workspace. On MemoryAgentBench histories longer than 256K tokens,
\sys achieves an overall score of 40.99\%, improving over
Compact+RAG (34.80\%), Compact-only (27.54\%), and Sliding Window
(17.95\%). On AgentLongBench, \sys maintains 53.0\% task success at
512K tokens and 50.0\% at 1M tokens. At 512K, its utility is comparable
to Compact+RAG (54.0\%), while at 1M it exceeds Compact+RAG by
8 percentage points and Compact-only by 16 points. These results show
that virtualizing the workspace allows useful historical state to
remain accessible even when the accumulated trajectory can no longer
fit within a single model invocation.

These results show that \sys preserves strong task utility across long-history workloads, including those beyond the model's native context window. We next examine whether this utility can be achieved efficiently. Unlike Compact+RAG, which must reinsert and prefill retrieved text, \sys directly restores reusable KV state. As we show next, this difference leads to a substantial gap in recovery efficiency.

\subsection{Efficiency}

As shown in Table~\ref{tab:utility-efficiency}, \sys maintains low pre-answer
latency across all evaluated workloads. Its latency ranges from
0.38\,s to 1.81\,s, whereas Compact+RAG requires 26.63--416.38\,s
when the cost of compaction is included. Even when summary-generation
time is excluded, Compact+RAG still requires 10.63--39.24\,s.
Across the five evaluated settings, \sys reduces this post-compaction
recovery latency by 11.4--53.8$\times$. Sliding Window remains cheaper
because it performs no historical recovery, but this efficiency comes
with the substantial utility loss.

The difference becomes particularly pronounced as the workspace grows.
On AgentLongBench, the pre-answer latency of \sys increases only from
0.38\,s ($\le$256K) to 0.62\,s (512K) and
0.73\,s (1M). In comparison, Compact+RAG increases from
111.39\,s to 263.70\,s and 416.38\,s, respectively. Even excluding
the cost of generating compact summaries, its recovery latency grows
from 14.80\,s to 23.58\,s and 39.24\,s. Thus, enlarging the
addressable workspace does not require \sys to reconstruct the
correspondingly larger history at each query; only the retrieved
working set needs to be restored into the bounded execution view.

Table~\ref{tab:longmemeval-utility-efficiency} provides a detailed breakdown
on LongMemEval-S. \sys processes the same 109.74K total historical
input tokens as Full Context, but only 0.08K tokens require fresh
prefill when answering the final query. Compact+RAG, in contrast,
processes 142.16K total input tokens and requires 31.00K tokens of
fresh prefill because retrieved historical text must be inserted into
the prompt and processed again. Consequently, \sys achieves a
pre-answer latency of only 0.48\,s, close to the 0.30\,s Full Context
reference, while Compact+RAG requires 26.63\,s in total.

These results show that preserving historical context as reusable KV
state fundamentally changes the cost of workspace recall: instead of
recomputing retrieved history through text prefill, \sys pays only the
cost of KV retrieval and restoration for the bounded working set. We
next examine whether this efficiency enables practical million-token
workspace virtualization on consumer hardware.

\subsection{Million-Token Workspace on a Consumer Device}

We next evaluate whether \sys (with \textsc{QW3}) can make million-token agent workspaces
practical under the tight memory budget of a consumer device. As
described in Section~6.1, all systems are evaluated on the same
off-the-shelf laptop with a 24\,GB RTX 5090 Laptop GPU using
Qwen3.6/3.8-27B NVFP4 and equivalent 8-bit KV-cache configurations.\footnote{
\sys and vLLM directly load the NVFP4 SafeTensors checkpoint and use
FP8 KV cache. Since llama.cpp operates on GGUF models and does not
support FP8 KV cache, we use the corresponding NVFP4 GGUF repack
(\url{https://huggingface.co/tngtech/Qwen3.6-27B-NVFP4-GGUF}) with
Q8\_0 KV cache. FP8 and Q8\_0 have the same KV-cache footprint in our
configuration, so this difference does not affect the GPU-memory
capacity comparison. Besides, Qwen3.6/3.8 27b have the same model architecture with the same memory footprint.}

Table~\ref{tab:consumer} compares the maximum context and workspace
capacities supported under this configuration. Conventional inference
engines remain constrained by GPU-resident KV capacity: vLLM supports
approximately 10K context tokens, while the latest llama.cpp
configuration we tested can accommodate approximately 80K tokens.
\sys supports the same 80K-token execution view as llama.cpp, but
decouples this active view from the addressable workspace, allowing the
workspace to grow to 1M tokens. This corresponds to four times the
model's native 256K-token context window and 12.5$\times$ the
GPU-resident execution view.

The three engines have different design priorities. vLLM is primarily designed for high-throughput serving and concurrent workloads, so its configuration is not specifically optimized for maximizing single-session context capacity. In contrast, llama.cpp places stronger emphasis on efficient local inference across resource-constrained hardware and therefore provides a more relevant reference for our single-session setting. Notably, QW3 matches llama.cpp with an 80K-token execution view under the same 24\,GB GPU budget, indicating that the underlying QW3 engine is already comparable in GPU-memory efficiency for local inference. KVMEM then builds on this efficient execution substrate to virtualize the addressable workspace from 80K to 1M tokens.

\begin{table*}[t]
\centering
\caption{Maximum context and workspace capacity on the same consumer
laptop with a 24\,GB RTX 5090 Laptop GPU.}
\label{tab:consumer}
\begin{tabular}{lccc}
\toprule
\textbf{System}
& \textbf{Execution View / Context}
& \textbf{Virtual Workspace}
& \textbf{Execution Model} \\
\midrule
vLLM      & 10K & --         & Direct context \\
llama.cpp & 80K & --         & Direct context \\
\sys (\textsc{QW3})      & 80K       & \textbf{1M} & Virtual workspace \\
\bottomrule
\end{tabular}
\end{table*}

Importantly, the large virtual workspace does not prevent responsive generation. Even with a 1M-token workspace and an 80K-token execution view, \sys sustains around 50 generated tokens/s on the consumer laptop in a single-session setting. Despite the tight 24\,GB GPU-memory budget, the consumer deployment therefore retains interactive generation speed while exposing a million-token workspace. These results show that KV-context virtualization can make million-token agent workspaces practical on consumer-grade hardware without requiring datacenter-class GPU memory.

\subsection{Workspace Scalability on the Server Platform}
\label{sec_exp:scalability}

We further evaluate how \sys scales as the addressable workspace grows
well beyond both GPU KV capacity and the model's native context window.
We conduct this experiment on the server platform and increase the
workspace size from 256K to 10M tokens while fixing the execution view
at 64K tokens. This isolates the overhead of enlarging the virtual
workspace from that of increasing the context processed by each model
invocation.

Table~\ref{tab:workspace_scalability} summarizes resource consumption and
execution performance across the evaluated workspace sizes. As the
workspace grows, the persistent KV repository and the attention-space
index expand, whereas the GPU memory footprint remains nearly constant
because the active execution view is fixed. Retrieval latency increases
with the number of addressable historical blocks, reflecting the larger
index that must be searched. Nevertheless, the resulting end-to-end
TTFT remains low, increasing only from 0.43\,s with a 256K-token
workspace to 1.60\,s with a 10M-token workspace. Decode throughput
remains stable across workspace sizes, while prefill throughput remains
on the order of 2K tokens/s even as the workspace scales to 10M tokens.
This shows expanding the addressable workspace primarily increases
retrieval and backing-storage costs while keeping the per-step execution
overhead practical.

\begin{table*}[t]
\centering
\caption{Workspace scalability on the server platform with 32-token blocks. The execution view is fixed at 64K tokens while the addressable workspace grows from 256K to 10M tokens. GPU Memory reports the total GPU-process footprint, Host Memory includes host-resident KV state and the retrieval index and is capped at 64 GiB, and NVMe Storage reports the persistent KV-store footprint. Retrieval latency and TTFT are medians over 16 frozen-query trials after one warm-up, using the same 128-token queries at all workspace sizes. Retrieval latency covers scoring, selection, and KV materialization, while TTFT additionally includes query prefill. Prefill throughput is measured with 2,048-token chunks.}
\label{tab:workspace_scalability}
\begin{tabular}{rccccccc}
\toprule
\textbf{Workspace}
& \textbf{GPU Memory}
& \textbf{Host Memory}
& \textbf{NVMe Storage}
& \textbf{Retrieval}
& \textbf{TTFT}
& \textbf{Prefill}
& \textbf{Decode} \\
& \textbf{(Model+KV, GiB)}
& \textbf{(KV+Index, GiB)}
& \textbf{(KV, GiB)}
& \textbf{Latency (s)}
& \textbf{(s)}
& \textbf{(tok/s)}
& \textbf{(tok/s)} \\
\midrule
256K & 33.9 & 18.0 &   8.5 & 0.174 & 0.427 & 2582.1 & 78.69 \\
512K & 34.7 & 26.5 &  17.0 & 0.209 & 0.466 & 2558.2 & 85.56 \\
1M   & 34.9 & 43.6 &  34.0 & 0.213 & 0.441 & 2639.3 & 83.25 \\
2M   & 34.8 & 64.0 &  68.0 & 0.393 & 0.625 & 2641.1 & 81.02 \\
4M   & 34.9 & 64.0 & 136.0 & 0.828 & 1.070 & 2597.5 & 75.67 \\
10M  & 34.9 & 64.0 & 324.2 & 1.311 & 1.601 & 1908.5 & 78.33 \\
\bottomrule
\end{tabular}
\end{table*}

The attention-space index is substantially more compact than the full
KV repository because Mean-K retrieval stores only one vector per
indexed attention layer and KV head for each historical block. Its
footprint grows from approximately 0.25 GiB with a 256K-token workspace to 9.5 GiB
with a 10M-token workspace. The index remains in host memory, with only bounded tiles staged on the GPU for scoring, so its growth does not increase the GPU memory used for retrieval.
Over the same range, median end-to-end retrieval latency, including scoring, selection, and KV materialization, increases from 174 ms to 1.311 s. Despite this growth, the latency remains on the order of one second even at a 10M-token workspace, making retrieval practical for interactive agent execution. 
memory and modest retrieval overhead.

Workspace expansion is absorbed primarily by the lower storage tiers.
Host-memory usage increases from 18\,GiB with a 256K-token workspace
to the configured limit of approximately 64\,GiB with a 2M-token
workspace and remains near this limit thereafter. As the workspace
continues to grow, additional historical KV state spills to the NVMe
backing store, whose footprint increases from 8.5\,GiB to 324\,GiB
across the evaluated range. In contrast, GPU memory remains
approximately 34\,GiB because the active execution view is fixed.
These results show that \sys scales workspace capacity by
absorbing growth in host memory and NVMe while keeping the GPU-memory
footprint bounded.

Overall, \sys scales the addressable workspace from 256K to 10M tokens while keeping the GPU-resident execution view fixed at 64K and maintaining practical model-execution performance. The 10M-token scale represents our evaluated upper bound rather than the capacity limit of the virtualization mechanism: additional workspace state can continue to spill to NVMe as backing-storage capacity increases. These results demonstrate the central benefit of workspace virtualization: the amount of historical state available to an agent can grow far beyond both GPU KV capacity and the model's native context window without requiring the entire workspace to reside on the GPU or be materialized into the active execution view at each step.

\subsection{Long-Horizon Agent Evaluation}
\label{sec:deepswe}

To test whether \sys remains effective during complete long-horizon
agent work, we evaluate it on DeepSWE v1.1, a repository-level
software-engineering benchmark with functional verifiers~\cite{huang2026deepswe}.
At the time of this experiment, the Qwen team had just released
Qwen3.8-27B, which specifically improves long-horizon agent execution.
We therefore use Qwen3.8-27B as the running model in this evaluation.
Unlike the controlled Qwen3.6-27B evaluations in previous sections,
which replay fixed long histories to isolate workspace-memory behavior,
this experiment evaluates \sys throughout complete agent trajectories,
where context management can affect subsequent reasoning, tool use,
and task outcomes.

We use the first sixteen tasks in the canonical DeepSWE v1.1 task
ordering. This subset is selected directly, without filtering tasks
according to their difficulty, trajectory length, or expected behavior
under either policy. Each task is sampled four times, allowing us to
measure both average rollout success and whether a policy can solve a
task in at least one of four attempts.

We compare \sys with the compaction-based method. Both methods use the
same Qwen3.8-27B-Q8 model, Claude Code harness, task prompts, initial
repository states, functional verifiers, sampling parameters, and set
of four random seeds per task. The \sys configuration uses a
1M-token logical workspace, a 128K-token selection budget, and a
64K-token generation reserve, with text compaction disabled. The
Compaction-only baseline instead uses a 256K-token dense context and
Claude Code's native auto-compaction. In total, we collect 64
trajectories for each configuration and 128 local trajectories overall.

\begin{table*}[t]
\centering
\small
\setlength{\tabcolsep}{8pt}
\renewcommand{\arraystretch}{1.10}
\caption{
Success on the first sixteen tasks in the canonical DeepSWE v1.1
task ordering. We select the first sixteen tasks directly, without
task-level filtering or cherry-picking, and sample each task four
times, yielding 64 trajectories per local configuration.
Public-model Pass@1 and Pass@4 values are reaggregated from released
scored trajectories over exactly the same task subset. Pass@4 denotes
the percentage of tasks with at least one passing scored rollout.
Public configurations use mini-swe-agent, whereas our paired
Qwen3.8-27B experiment uses Claude Code.
}
\label{tab:deepswe-success}
\begin{tabular}{@{}llccc@{}}
\toprule
\textbf{Model and configuration}
& \textbf{Agent harness}
& \textbf{Runs/task}
& \textbf{Pass@1 (\%) $\uparrow$}
& \textbf{Pass@4 (\%) $\uparrow$} \\
\midrule
\multicolumn{5}{@{}l}{\textit{Public DeepSWE trajectories}} \\
Gemini 3.7 Flash [high]
& mini-swe-agent & 4 & \textbf{64.1} & \underline{81.3} \\
DeepSeek V4 Flash [max]
& mini-swe-agent & 4 & \underline{53.1} & 75.0 \\
Claude Sonnet 4.6 [high]
& mini-swe-agent & 4 & 32.8 & 56.3 \\
Kimi K2.7 Code [default]
& mini-swe-agent & 4 & 29.7 & 75.0 \\
\midrule
\multicolumn{5}{@{}l}{%
\textit{Our paired Qwen3.8-27B experiment}} \\
Qwen3.8-27B + \sys
& Claude Code & 4 & 48.4 & \textbf{93.8} \\
Qwen3.8-27B + Compaction-only
& Claude Code & 4 & 43.8 & \underline{81.3} \\
\bottomrule
\end{tabular}
\end{table*}

Table~\ref{tab:deepswe-success} reports the task-success results.
Across the 64 rollouts for each configuration, \sys achieves a
Pass@1 of 48.4\%, compared with 43.8\% for the Compaction-only
baseline. This corresponds to 31 versus 28 successful rollouts and an
absolute improvement of 4.7 percentage points. The improvement is more
pronounced at the task level: \sys increases Pass@4 from 81.3\% to
93.8\%, a gain of 12.5 percentage points. In other words, \sys solves
15 of the 16 tasks in at least one of four runs, whereas the
Compaction-only baseline solves 13. These results indicate that
workspace virtualization improves both average rollout reliability and
the breadth of tasks that the model can solve.

Table~\ref{tab:deepswe-success} also reports results reaggregated from
publicly released DeepSWE trajectories over exactly the same sixteen
tasks. The 48.4\% Pass@1 of Qwen3.8-27B with \sys is below Gemini
3.7 Flash and DeepSeek V4 Flash, but above the reported Claude Sonnet
4.6 and Kimi K2.7 Code configurations. More notably, \sys achieves the
highest Pass@4 in the table, at 93.8\%. This comparison suggests that
the Qwen3.8-27B configuration with \sys provides strong task-level
coverage, although the public configurations use mini-swe-agent rather
than Claude Code and therefore do not constitute a strictly controlled
model comparison.

\begin{table*}[t]
\centering
\small
\setlength{\tabcolsep}{8pt}
\renewcommand{\arraystretch}{1.10}
\caption{
Efficiency on the first 16 DeepSWE tasks.
Compared with the compaction-only Qwen3.8-27B baseline, \sys achieves
a 1.15$\times$ agent speedup, a 1.18$\times$ request-time speedup,
a 2.23$\times$ prefill speedup, and a 1.23$\times$ reduction in
decoded tokens.
Values are per-task averages, and parenthetical factors report the
Compaction-only value divided by the \sys value.
}
\label{tab:deepswe-efficiency}

\begin{tabular}{@{}llcccc@{}}
\toprule
\textbf{Sample group}
& \textbf{Policy}
& \makecell{\textbf{Agent}\\\textbf{time (min)}}
& \makecell{\textbf{Request}\\\textbf{wall time (min)}}
& \makecell{\textbf{Prefill}\\\textbf{time (s)}}
& \makecell{\textbf{Decoded}\\\textbf{tokens (K)}} \\
\midrule

\multirow{2}{*}{\makecell[l]{All selected tasks\\($n=16$)}}
& \sys
& \textbf{45.8} (1.15$\times$)
& \textbf{40.7} (1.18$\times$)
& \textbf{95.0} (2.23$\times$)
& \textbf{125.6} (1.23$\times$) \\

& Compaction-only
& 52.5
& 48.1
& 211.5
& 154.4 \\

\bottomrule
\end{tabular}
\end{table*}

Table~\ref{tab:deepswe-efficiency} shows the efficiency results over
all sixteen tasks. The largest improvement is in prefill processing:
\sys reduces the average prefill time from 211.5 to 95.0 seconds, a
55.1\% reduction and a 2.23$\times$ speedup. These savings translate
into lower end-to-end execution time. Average agent time decreases from
52.5 to 45.8 minutes, corresponding to a 12.8\% reduction and a
1.15$\times$ speedup, while request wall time decreases from 48.1 to
40.7 minutes, corresponding to a 15.4\% reduction and a
1.18$\times$ speedup. \sys also reduces the average decoded output
from 154.4K to 125.6K tokens, an 18.7\% reduction, with a
Compaction-only-to-\sys ratio of 1.23$\times$. Thus, the reduction in
workspace-processing overhead does not merely shift computation to
longer model outputs; \sys reduces both input-processing time and
decoded-token consumption.

Taken together, these results show that the utility and efficiency
advantages of \sys extend from controlled long-history workloads to
complete software-engineering trajectories. Across four rollouts on
each of sixteen tasks, \sys improves Pass@1 by 4.7 percentage points
and Pass@4 by 12.5 percentage points, solving 15 of the 16 tasks in at
least one run. At the same time, it reduces all reported efficiency
metrics, including a 55.1\% reduction in prefill time, a 15.4\%
reduction in request wall time, and a 12.8\% reduction in end-to-end
agent time. These findings indicate that preserving a large logical
workspace while selecting a bounded active context can improve both
the effectiveness and execution efficiency of long-running agents,
relative to repeatedly compressing their accumulated histories.

\section{Discussion and Limitations}
\label{sec:discussion}

\noindent \textbf{Scope of Workspace Virtualization.}
\sys virtualizes the historical workspace available to an agent; it
does not extend the number of tokens jointly attended to in a single
model invocation. Each step still operates on a bounded execution view
limited by the model context window and GPU KV capacity. The 1M- and
10M-token results therefore refer to addressable virtual workspaces,
from which only query-relevant state is materialized at each step.

\smallskip

\noindent \textbf{Fidelity of KV Reuse.}
Reusing historical KV state is not mathematically equivalent to
recomputing the corresponding text under the newly assembled context.
\sys restores positional consistency through re-RoPE and re-prefills
the current query over the updated execution view, but historical KV
was originally computed under its earlier causal context.
Our evaluation shows that reused KV state preserves strong utility on the studied workloads. In addition, because each step materializes only a retrieved subset of the workspace, task-relevant blocks may still be missed by retrieval. When resources permit, longer-workspace tasks may therefore benefit from a larger active execution view, reducing reliance on retrieval and the risk of excluding relevant context.

\smallskip

\noindent \textbf{Storage Overhead.}
\sys trades backing-storage capacity for reusable model state: KV state
is substantially larger than text, and larger workspaces increase
host-memory/NVMe usage, retrieval-index size, and retrieval overhead.
The 10M-token workspace is the largest scale evaluated in this work,
rather than a hard architectural limit; larger workspaces can be
supported with additional backing storage at increased storage and
retrieval cost.

\smallskip

\noindent \textbf{Deployment Scope.}
Our current implementation requires control over KV allocation,
retrieval, positional restoration, and tier movement, and therefore
cannot be transparently layered over black-box LLM APIs. Integrating
workspace virtualization into cloud serving stacks is a promising
direction: providers could expose larger persistent agent workspaces
while reducing repeated prefill, potentially improving serving
economics given the large price gap between freshly processed and
cache-reused input in existing APIs~\cite{deepseek_pricing}. Extending
\sys to concurrent multi-user serving, where multiple workspaces
compete for shared GPU and storage resources, is another direction for
future work.
\section{Conclusion}
\label{sec:conclusion}

We present \sys, a KV-context virtualization system that decouples an
agent's addressable workspace from the KV state physically resident on
the GPU. Instead of compacting overflowed history into lossy text,
\sys preserves previously processed workspace state as paged KV across
GPU memory, host memory, and NVMe, and recalls a bounded,
query-dependent execution view at each agent step. Step-level memory
scheduling, query-conditioned KV retrieval, and tiered KV management
make this virtualization practical while avoiding repeated prefill of
historical text.

Across three long-context agent benchmarks, \sys consistently
outperforms sliding-window and compact-only context management and
generally matches or exceeds retrieval-augmented compaction, while
reducing post-compaction recovery latency by 11.4--53.8$\times$. In a paired 16-task DeepSWE evaluation with Qwen3.8-27B,
\sys increases Pass@1 from 43.8\% to 48.4\% and Pass@4
from 81.3\% to 93.8\%.
With Qwen3.6/3.8-27B, \sys makes million-token agent workspaces practical
on consumer hardware: on a laptop with a 24\,GB RTX 5090 Laptop GPU,
it supports a 1M-token virtual workspace with an 80K-token execution
view while sustaining approximately 50 tokens/s in single-session
generation. On our server platform, \sys further scales the
Qwen3.6-27B workspace to 10M tokens while keeping the execution view
bounded. These results show that long-running agent workspaces can grow
far beyond both GPU KV capacity and the model's native context window,
providing a practical path toward substantially larger persistent
workspaces on commodity hardware.

\balance
\bibliographystyle{plain}
\bibliography{references}

@misc{anthropic2026contextwindow,
  author = {Anthropic},
  title = {{Claude Code} context window},
  howpublished = {\url{https://code.claude.com/docs/en/context-window}},
  note = {Accessed June 2026},
  year = {2026}
}

@misc{anthropic2026memory,
  author = {Anthropic},
  title = {How {Claude} remembers your project},
  howpublished = {\url{https://code.claude.com/docs/en/memory}},
  note = {Accessed June 2026},
  year = {2026}
}

@article{chhikara2025mem0,
  author = {Chhikara, Prateek and Khant, Dev and Aryan, Saket and Singh, Taranjeet and Yadav, Deshraj},
  title = {{Mem0}: Building Production-Ready {AI} Agents with Scalable Long-Term Memory},
  journal = {arXiv preprint arXiv:2504.19413},
  year = {2025}
}

@article{cheng2025lmcache,
  author = {Cheng, Yihua and Liu, Yuhan and Li, Hanchen and Yao, Jiayi and Ray, Siddhant and others},
  title = {{LMCache}: An Efficient {KV} Cache Layer for Enterprise-Scale {LLM} Serving},
  journal = {arXiv preprint arXiv:2510.09665},
  year = {2025}
}

@article{kang2025memoryos,
  author = {Kang, Jiazheng and Ji, Mingming and Zhao, Zhe and Bai, Ting},
  title = {Memory {OS} of {AI} Agent},
  journal = {arXiv preprint arXiv:2506.06326},
  year = {2025}
}

@inproceedings{kwon2023vllm,
  author = {Kwon, Woosuk and Li, Zhuohan and Zhuang, Siyuan and Sheng, Ying and Zheng, Lianmin and Yu, Cody Hao and Gonzalez, Joseph E. and Zhang, Hao and Stoica, Ion},
  title = {Efficient Memory Management for Large Language Model Serving with PagedAttention},
  booktitle = {Proceedings of the 29th Symposium on Operating Systems Principles (SOSP)},
  year = {2023}
}

@article{liu2024lost,
  author = {Liu, Nelson F. and Lin, Kevin and Hewitt, John and Paranjape, Ashwin and Bevilacqua, Michele and Petroni, Fabio and Liang, Percy},
  title = {Lost in the Middle: How Language Models Use Long Contexts},
  journal = {Transactions of the Association for Computational Linguistics},
  volume = {12},
  pages = {157--173},
  year = {2024}
}

@inproceedings{wu2025longmemeval,
  author = {Wu, Di and Wang, Hongwei and Yu, Wenhao and Zhang, Yuwei and Chang, Kai-Wei and Yu, Dong},
  title = {{LongMemEval}: Benchmarking Chat Assistants on Long-Term Interactive Memory},
  booktitle = {Proceedings of the International Conference on Learning Representations (ICLR)},
  year = {2025}
}

@article{nan2025nemori,
  author = {Nan, Jiayan and Ma, Wenquan and Wu, Wenlong and Chen, Yize},
  title = {{Nemori}: Self-Organizing Agent Memory Inspired by Cognitive Science},
  journal = {arXiv preprint arXiv:2508.03341},
  year = {2025}
}

@misc{openai2025codex,
  author = {OpenAI},
  title = {Introducing {Codex}},
  howpublished = {\url{https://openai.com/index/introducing-codex/}},
  note = {Accessed June 2026},
  year = {2025}
}

@misc{openclaw2026compaction,
  author = {OpenClaw},
  title = {Compaction},
  howpublished = {\url{https://docs.openclaw.ai/concepts/compaction}},
  note = {Accessed June 2026},
  year = {2026}
}

@misc{openclaw2026context,
  author = {OpenClaw},
  title = {Context},
  howpublished = {\url{https://docs.openclaw.ai/concepts/context}},
  note = {Accessed June 2026},
  year = {2026}
}

@misc{openhands2026contextcondenser,
  author = {OpenHands},
  title = {Context condenser},
  howpublished = {\url{https://docs.openhands.dev/sdk/guides/context-condenser}},
  note = {Accessed June 2026},
  year = {2026}
}

@misc{openhands2026workspace,
  author = {OpenHands},
  title = {Workspace architecture},
  howpublished = {\url{https://docs.openhands.dev/sdk/arch/workspace}},
  note = {Accessed June 2026},
  year = {2026}
}

@article{packer2023memgpt,
  author = {Packer, Charles and Fang, Vivian and Patil, Shishir G. and Lin, Kevin and Wooders, Sarah and Gonzalez, Joseph E.},
  title = {{MemGPT}: Towards {LLM}s as Operating Systems},
  journal = {arXiv preprint arXiv:2310.08560},
  year = {2023}
}

@inproceedings{xiao2024streamingllm,
  title={Efficient streaming language models with attention sinks},
  author={Xiao, Guangxuan and Tian, Yuandong and Chen, Beidi and Han, Song and Lewis, Mike},
  booktitle={International Conference on Learning Representations},
  volume={2024},
  pages={21875--21895},
  year={2024}
}

@article{yang2024sweagent,
  author = {Yang, John and Jimenez, Carlos E. and Wettig, Alexander and Lieret, Kilian and Yao, Shunyu and Narasimhan, Karthik and Press, Ofir},
  title = {{SWE-agent}: Agent-Computer Interfaces Enable Automated Software Engineering},
  journal = {arXiv preprint arXiv:2405.15793},
  year = {2024}
}

@inproceedings{yao2025cacheblend,
  author = {Yao, Jiayi and Li, Hanchen and Liu, Yuhan and Ray, Siddhant and Cheng, Yihua and Zhang, Qizheng and Du, Kuntai and Lu, Shan and Jiang, Junchen},
  title = {{CacheBlend}: Fast Large Language Model Serving for {RAG} with Cached Knowledge Fusion},
  booktitle = {Proceedings of the Twentieth European Conference on Computer Systems (EuroSys)},
  year = {2025}
}

@article{zhang2023h2o,
  title = {{H2O}: Heavy-Hitter Oracle for Efficient Generative Inference of Large Language Models},
  author = {Zhang, Zhenyu and Sheng, Ying and Zhou, Tianyi and Chen, Tianlong and Zheng, Lianmin and Cai, Ruisi and Song, Zhao and Tian, Yuandong and R{\'e}, Christopher and Barrett, Clark and others},
  journal = {Advances in Neural Information Processing Systems},
  volume = {36},
  pages = {34661--34710},
  year = {2023}
}

@inproceedings{zheng2024sglang,
  author = {Zheng, Lianmin and Yin, Liangsheng and Xie, Zhiqiang and Sun, Chuyue and Huang, Jeff and Yu, Cody Hao and Cao, Shiyi and Kozyrakis, Christos and Stoica, Ion and Gonzalez, Joseph E. and Barrett, Clark and Sheng, Ying},
  title = {{SGLang}: Efficient Execution of Structured Language Model Programs},
  booktitle = {Advances in Neural Information Processing Systems},
  year = {2024}
}

@misc{fang2026agentlongbench,
title = {AgentLongBench: A Controllable Long Benchmark For Long-Contexts Agents via Environment Rollouts},
author = {Fang, Shicheng and Wang, Yuxin and Liu, Xiaoran and Lu, Jiahao and Tan, Chuanyuan and Chen, Xinchi and Zheng, Yining and Huang, Xuanjing and Qiu, Xipeng},
year = {2026},
eprint = {2601.20730},
archivePrefix = {arXiv},
primaryClass = {cs.CL}
}

@misc{hu2025memoryagentbench,
title         = {Evaluating Memory in LLM Agents via Incremental Multi-Turn Interactions},
author        = {Hu, Yuanzhe and Wang, Yu and McAuley, Julian},
year          = {2025},
eprint        = {2507.05257},
archivePrefix = {arXiv},
primaryClass  = {cs.CL}
}

@article{hsieh2024ruler,
  title={RULER: What's the real context size of your long-context language models?},
  author={Hsieh, Cheng-Ping and Sun, Simeng and Kriman, Samuel and Acharya, Shantanu and Rekesh, Dima and Jia, Fei and Zhang, Yang and Ginsburg, Boris},
  journal={arXiv preprint arXiv:2404.06654},
  year={2024}
}

@inproceedings{pan2024llmlingua,
  title={Llmlingua-2: Data distillation for efficient and faithful task-agnostic prompt compression},
  author={Pan, Zhuoshi and Wu, Qianhui and Jiang, Huiqiang and Xia, Menglin and Luo, Xufang and Zhang, Jue and Lin, Qingwei and R{\"u}hle, Victor and Yang, Yuqing and Lin, Chin-Yew and others},
  booktitle={Findings of the Association for Computational Linguistics: ACL 2024},
  pages={963--981},
  year={2024}
}

@misc{deepseek_pricing,
  author       = {{DeepSeek}},
  title        = {DeepSeek API Models and Pricing},
  year         = {2026},
  howpublished = {\url{https://api-docs.deepseek.com/quick_start/pricing/}},
  note         = {Accessed August 13, 2026}
}

@article{huang2026deepswe,
  title   = {{DeepSWE}: Measuring Frontier Coding Agents on Original,
             Long-Horizon Engineering Tasks},
  author  = {Huang, Wenqi and Lee, Charley and Tng, Leonard and Ge, Serena},
  journal = {arXiv preprint arXiv:2607.07946},
  year    = {2026},
  url     = {https://arxiv.org/abs/2607.07946}
}

\end{document}